\documentclass[10pt,twocolumn,letterpaper]{article}

\usepackage{cvpr}
\usepackage{times}
\usepackage{epsfig}
\usepackage{graphicx}
\usepackage{amsmath}
\usepackage{amssymb}
\usepackage{bbm}

\usepackage[utf8]{inputenc}
\usepackage{geometry}
\usepackage{subcaption}
\usepackage[justification=centering]{caption}
\usepackage{colortbl}
\definecolor{lightgray}{rgb}{0.9,0.9,0.9}

\usepackage[breaklinks=true,bookmarks=false]{hyperref}

\cvprfinalcopy 

\def\cvprPaperID{****} 

\begin{document}

\title{DeepFace: Face Generation using Deep Learning}
\author{Hardie Cate\\
{\tt\small ccate@stanford.edu}
\and
Fahim Dalvi\\
{\tt\small fdalvi@cs.stanford.edu}
\and
Zeshan Hussain\\
{\tt\small zeshanmh@stanford.edu}
}

\maketitle

\begin{abstract}
We use CNNs to build a system that both classifies images of faces based on a variety of different facial attributes and generates new faces given a set of desired facial characteristics. 
After introducing the problem and providing context in the first section, we discuss recent work related to image generation in Section 2.
In Section 3, we describe the methods used to fine-tune our CNN and generate new images using a novel approach inspired by a Gaussian mixture model.
In Section 4, we discuss our working dataset and describe our preprocessing steps and handling of facial attributes.
Finally, in Sections 5, 6 and 7, we explain our experiments and results and conclude in the following section.
Our classification system has 82\% test accuracy.
Furthermore, our generation pipeline successfully creates well-formed faces.
\end{abstract}

\section{Introduction}
Convolutional neural networks (CNNs) are powerful tools for image classification and object detection, but they can also be used to generate images.
There are myriad image generation tasks and applications, ranging from generating potential mass lesions in radiology scans to creating landscapes or artistic scenes.
In our project, we address the problem of generating faces based on certain desired facial characteristics.
Potential facial characteristics fall within the general categories of raw attributes (e.g., \textit{big nose}, \textit{brown hair}, etc.), ethnicity (e.g., \textit{white}, \textit{black}, \textit{Indian}), and accessories (e.g. \textit{sunglasses}, \textit{hat}, etc.). 
The problem can be stated as follows:
Given a set of facial attributes as input, produce an image of a well-formed face as output that contains these characteristics.

In our face generation system, we fine-tune a CNN pre-trained on faces to create a classification system for facial characteristics. 
We employ a novel technique that models distributions of feature activations within the CNN as a customized Gaussian mixture model.
We then perform feature inversion using the relevant features identified by this model.
The primary advantage of our implementation is that it does not require any deep learning architectures apart from a CNN whereas other generative approaches do.
Our face generation system has many potential uses, including identifying suspects in law enforcement settings as well as in other more generic generative settings.

\section{Related Work}
Work surrounding generative models for deep learning has mostly been in developing graphical models, autoencoder frameworks, and more recently, generative recurrent neural networks (RNNs). Specific graphical models that have been used to learn generative models of data are Restricted Boltzmann Machines (RBMs), an undirected graphical model with connected stochastic visible and stochastic hidden units, and their generalizations, such as Gaussian RBMs. Srivastava and Salakhutdinov use these basic RBMs to create a Deep Boltzmann Machine (DBM), a multimodal model that learns a probability density over the space of multimodal inputs and can be effectively used for information retrieval and classification tasks \cite{NIPS20124683}. Similar work done by Salakhutdinov and Hinton shows how the learning of a high capacity DBM with multiple hidden layers and millions of parameters can be made more efficient with a layer-by-layer "pre-training" phase that allows for more reasonable weight initializations by incorporating a bottom-up pass \cite{salakhutdinov2009deep}. In his thesis, Salakhutdinov also adds to this learning algorithm by incorporating a top-down feedback pass as well as a bottom-up pass, which allows DBMs to better propagate uncertainty about ambiguous inputs \cite{salakhutdinov2009learning}.

Other generative approaches involve using autoencoders. The first ideas regarding the probabilistic interpretation of autoencoders were proposed by Ranzato et al.; a more formal interpretation was given by Vincent, who described denoising autoencoders (DAEs) \cite{marc2007efficient} \cite{vincent2011connection}. A DAE takes an input $\mathbf{x} \in [0,1]^d$ and first maps it, with an encoder, to a hidden representation $\mathbf{y} \in [0,1]^{d'}$ through some mapping, $\mathbf{y} = s(\mathbf{W}\mathbf{x} + \mathbf{b})$, where $s$ is a non-linearity such as a sigmoid. The latent representation $\mathbf{y}$ is then mapped back via a decoder into a reconstruction $\mathbf{z}$ of the same shape as $\mathbf{x}$, i.e. $\mathbf{z} = d(\mathbf{W}'\mathbf{y} + \mathbf{b}')$. The parameters, $\mathbf{W}$, $\mathbf{b}$, $\mathbf{b}'$, and $\mathbf{W}'$ are learned such that the average reconstruction loss between $\mathbf{x}$ and $\mathbf{z}$ is minimized \cite{deep2016tutorial}. Bengio et al. show an alternate form of the DAE: given some observed input $X$ and corrupted input $\widetilde{X}$, where $\widetilde{X}$ has been corrupted based on a conditional distribution $C(\widetilde{X}|X)$, we train the DAE to estimate the reverse conditional $P(X|\widetilde{X})$ \cite{bengio2013generalized}. With this formulation, Vincent et al. construct a deeper network of stacked DAEs to learn useful representations of the inputs \cite{vincent2010stacked}.

An alternate model has been posited by Gregor et al., who propose using a recurrent neural network architecture to generate digits. This architecture is a type of variational autoencoder, a recent advanced model that bridges deep learning and variational inference, since it is comprised of an encoder RNN that compresses the real images during training and a decoder RNN that reconstitutes images after receiving codes \cite{gregor2015draw}. 

Finally, another approach for estimating generative models is via generative adversarial nets \cite{gauthier2014conditional} \cite{goodfellow2014generative}. In this framework, two models are simultaneously trained: a generative model $G$ that captures the distribution of the data and a discriminative model $D$ that estimates the probability that a sample came from the training data rather than $G$. $G$ is trained to maximize the probability that $D$ makes a mistake. 

\section{Methods}
\begin{figure*}
  \includegraphics[width=\textwidth,height=5cm]{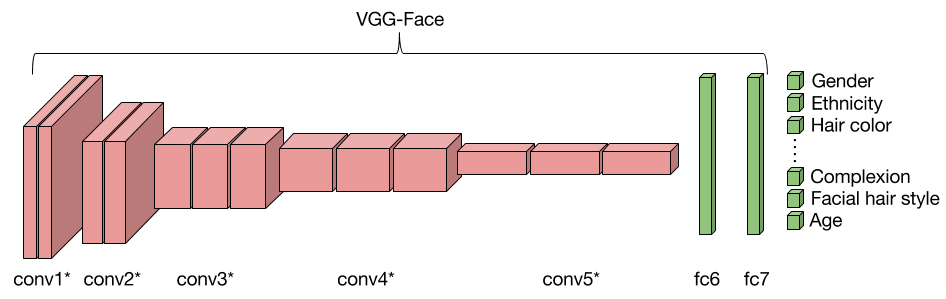}
  \caption{Modified VGG-Face Architecture}
  \label{fig:arch}
\end{figure*}
Our workflow consists of two steps. Firstly, we finetune a pre-trained model to classify facial and image characteristics for an input image. Secondly, we generate faces given some description. We describe our approach for each step in the forthcoming sections.

\subsection{Notation}
To facilitate the discussion of our methods, we introduce some notation. Let $D$ denote our set of training images, and let $F$ denote the set of 73 facial attributes.
From this point forward, we use the word \textit{attribute} to refer strictly to one of the 73 facial characteristics in our system.

\subsection{Fine-tuning}\label{sec:finetuningmethod}
For finetuning, we employ the VGG-Face net, a 16-layer CNN that was trained on ~2 million celebrity faces and evaluated on faces from the Labeled Faces in the Wild and YouTube faces datasets \cite{vggfacenet}. Using the VGG-Face net as our base architecture, we then attach $42$ heads to the end of the fc-7 layer of the VGG-Face net, each of which consists of a fully connected fc-8 layer and a softmax layer. Each softmax head is a multilabel classifier for a group of attributes that are highly correlated. For example, one group of attributes includes hair color (\textit{black}, \textit{gray}, \textit{brown}, and \textit{blond}); in general, someone usually only has one color of hair, which makes grouping these particular attributes together as one multilabel classification reasonable. Furthermore, grouping together highly correlated attributes for our classification task in lieu of having $73$ binary classifiers provides the network with implicit information on the relationship between grouped features.

During training, we freeze certain layers in the CNN and learn the weights and biases on the unfrozen layers. To determine which layers to freeze, we run several experiments with different sets of frozen layers and choose the set with the optimal performance. We describe the experiments to assess our architecture and the results of the fine-tuning in Section 5.  


\subsection{Generation}
\subsubsection{Baseline Approach}
We begin the generation phase of our project with a simple baseline approach that uses class visualization.
This approach begins with the mean image $M$, which is the pixel-wise and channel-wise mean of all images in our training set. We add small Gaussian noise to $M$ and use this image $X$ as our input to our baseline algorithm. Suppose we wish to produce a face with a set of attributes $C\subseteq F$. In each iteration of our algorithm, we do the following: 
    \begin{enumerate}
        \item Perform a forward pass with $X$ as input;
        \item Set the gradients at each softmax layer to either 1 or 0, depending on our target attributes $C$;
        \item Perform a backward pass through the network with these gradients;
        \item Update $X$ using stochastic gradient descent (SGD) and regularization.
    \end{enumerate}
This baseline serves primarily to demonstrate that our CNN correctly represents important facial structures such as eyes, noses, mouths, etc., even after fine-tuning on specific facial attributes.
This approach is not comprehensive enough because boosting certain features from the last layer in the network does not limit the number of facial structures that appear in the image.
A more complete discussion of the results of the baseline approach can be found in the Section 5,6.

Our next and final approach uses a customized variant of a Gaussian Mixture Model (GMM). We opt for this technique over more traditional generative approaches discussed in Section 2 for two main reasons. 
Firstly, to our knowledge, this method is novel and has not been used in context of image generation using CNNs.
Secondly, this approach does not require complex structures in addition to our existing CNN, so the model is simpler and has far fewer parameters to train.

\subsubsection{Custom Gaussian Mixture Model}
This approach behaves similarly to the baseline in that we begin with an input image $X$ that consists of random noise and boost this image toward a set of attributes $C\subseteq F$.
The primary difference is that instead of class visualization, we use feature inversion with respect to a layer in the CNN. We also begin with random noise instead of the mean image.

Intuitively, feature inversion tries to minimize the difference in the activations of an input image and some target activations in a certain layer of the CNN.
These target activations are the activations that we believe an input image with the desired attributes $C$ will have when passed through the CNN.

More formally, Let $\phi_l(I)$ be the activations at layer $l$ when an image $I$ is passed through the CNN. 
Let $T_l$ denote the target activations for layer $l$ and suppose we have a way of determining $T_l$ from $C$.
We wish to find an image $I^*$ by solving the optimization problem
\begin{equation}
    I^* = \arg\min_{I} \Vert T_l - \phi_l(I) \Vert_2^2 + R(I)
\end{equation}
where $\Vert \cdot \Vert_2^2$ is the squared Euclidean norm and $R$ is a regularizer such as blurring and/or jitter.
Given $T_l$ and $X$, feature inversion does the following:
    \begin{enumerate}
        \item Perform a forward pass with $X$ as input up to layer $l$;
        \item Set the gradients at $l$ by taking the gradient of the objective function above with respect to the activations in layer $l$;
        \item Perform a backward pass from $l$ with these gradients;
        \item Update $X$ using SGD and regularization.
    \end{enumerate}

The challenge of this approach is determining an appropriate set of target activations $T_l$ given $C$.
To our knowledge, there is no established way to automatically detect these target activations given solely the desired facial attributes.

To address this problem, we introduce a custom Gaussian Mixture Model (cGMM), in which we model the distribution of $T_l$ for a given attribute $f$ as a multivariate Gaussian distribution.
For each $f\in F$, we estimate the mean and covariance matrix of its Gaussian by sampling images in our training set that are positive examples for this attribute.
For computational simplicity, we assume the covariance matrix for each distribution is diagonal.
This assumption implies that we effectively treat each multivariate Gaussian as a stacking of many independent Gaussians that model the distribution of activations for a single neuron in $l$.

Specifically, for each attribute $f_1, f_2, \ldots, f_{73}\in F$, we select 73 random sets of images $S_1,S_2, \ldots, S_{73}\subseteq D$ of equal size such that for each $i$, every image in $S_i$ has attribute $f_i$.
Then for each $i$, we compute a mean vector $\mu_i$ and covariance matrix $\Sigma_i$ based on the activations at layer $l$ of the images in $S_i$.
These mean vectors and covariance matrices define 73 independent multivariate Gaussian distributions.

Thus, according to our model, if we wish to produce possible activations at $l$ of an image that has attribute $f_i$ and no other $f\in F$, we can simply sample from $\mathcal{N}(\mu_i, \Sigma_i)$.
More generally, we can estimate the target activations for a set of attributes by performing a weighted sum of samples from the Gaussians, which we assume to be independent. 
If we wish to estimate the target activations at $l$ for a subset $C\subseteq F$ of attributes, we would produce a sample $s$ given by
\begin{equation}
    s = \frac{1}{|C|}\sum_{i=1}^{73} \mathbbm{1}_i w_i z_i
\end{equation}
where $\mathbbm{1}_i=1$ if $f_i\in C$ and 0 otherwise, $w_i$ is the weight assigned to the $i$th Gaussian, and $z_i\sim \mathcal{N}(\mu_i, \Sigma_i)$ is a random variable.
This $s$ can be used in place of $T_l$ for feature inversion.

Now we must determine the weights vector $w = (w_1, \ldots, w_{73})$.
In our approach, we learn this weights vector by trying to find weights that minimize the difference between the target activations produced by the sampling method above and the actual activations for images.
We wish to find these weights by solving the optimization problem
\begin{equation}
    w^* = \arg \min_w \frac{1}{|D|}\sum_{i=1}^{|D|} \Vert \phi_l(I_i) - \Phi \Vert_2^2 + \lambda\Vert w \Vert_2^2
\end{equation}
where
\begin{equation}
    \Phi = \frac{1}{|C_i|}\sum_{j=1}^{73} \mathbbm{1}_{i,j} w_j \mu_j
\end{equation}

and $\mathbbm{1}_{i,j}$ indicates whether $f_j\in C_i$, the attribute set for image $i$.
We learn the weights by taking derivatives of this objective function with respect to $w$ and iteratively updating $w$ using SGD.
\section{Datasets and Features}
\begin{figure}[h]
    \centering
    \begin{subfigure}[b]{0.24\linewidth}
        \includegraphics[width=\linewidth]{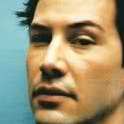}
        \label{fig:keanu}
    \end{subfigure}
    \hfill
    \begin{subfigure}[b]{0.24\linewidth}
        \includegraphics[width=\linewidth]{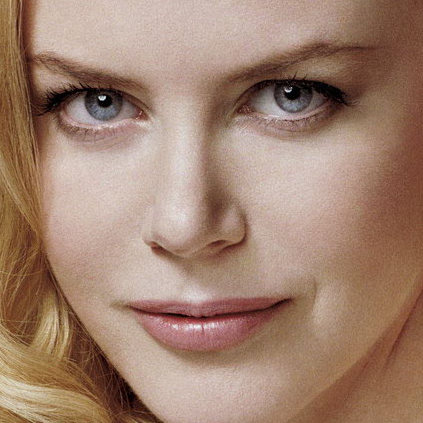}
        \label{fig:nicole}
    \end{subfigure}
    \hfill
    \begin{subfigure}[b]{0.24\linewidth}
        \includegraphics[width=\linewidth]{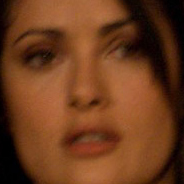}
        \label{fig:morgan}
    \end{subfigure}
    \hfill
    \begin{subfigure}[b]{0.24\linewidth}
        \includegraphics[width=\linewidth]{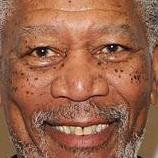}
        \label{fig:morgan}
    \end{subfigure}
    \caption{Variations in lighting conditions and camera angles}\label{fig:dataset}
\end{figure}
We are currently using the PubFig dataset \cite{dataset}, which is a collection of faces from public photos of 200 individuals. The dataset includes 58K images scraped from various sources on the Internet, and bounding boxes around the face for each of these images. However, to avoid copyright issues, the authors provide links to the images, instead of hosting the images themselves. We have roughly $20,000$ images in our training set and $8,000$ images in our test set. The missing images include those that do not exist at the given links, those that have changed and a few that had incorrect bounding boxes for the faces. 

Since these images are scraped from public photographs, there are several variations in faces. Most faces are frontal, while some are at slight angles. There is also a wide spectrum of lighting conditions in the dataset. 

For each image in the dataset, we also have a list of 73 attributes including gender, eye color, face shape and ethnicity. Many of these attributes are interdependent. For example, only one of the attributes within \textit{black hair}, \emph{blond hair} and \emph{brown hair} is positively labeled for each individual.

\section{Experiments \& Analysis}

\subsection{Finetuning Experiments}
\begin{figure*}
  \includegraphics[width=\textwidth,height=4cm]{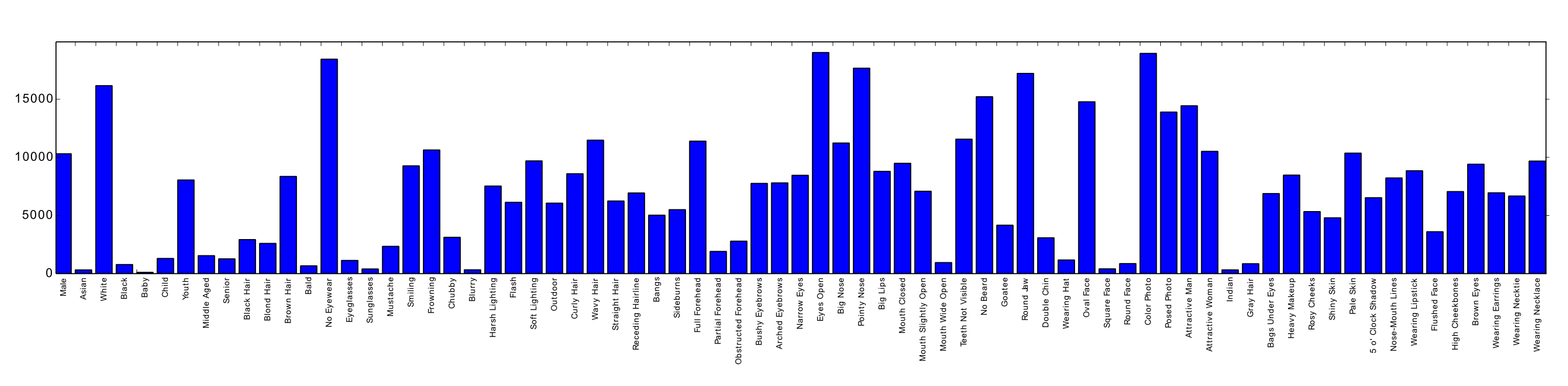}
  \caption{Distribution of all attributes in our training dataset}
  \label{fig:attrib-distrib}
\end{figure*}

To understand the data better, we first run some experiments to find out what the distribution of our data looks like. For each image, we are given information about the presence or absence of 73 attributes. In Figure \ref{fig:attrib-distrib} we see that the distribution is quite skewed towards certain attributes. For example, around $76\%$ of the people in the dataset are white, while only $3\%$ are black. The number of people of other ethnicities is even lower, so a large portion of the images do not have any ethnicity labeling. Another prime example of such a disparity is eyewear. Around $90\%$ of the people in the dataset have no form of eyewear in the photos. As another example of the incompleteness of our dataset, only $60\%$ of the people have age information associated with their faces. These numbers give us a good idea of the limits of the system we are building and how certain attributes might generate better faces than others. 
\begin{figure}
    \centering
    \hfill
    \begin{subfigure}[b]{0.49\linewidth}
        \includegraphics[width=\linewidth]{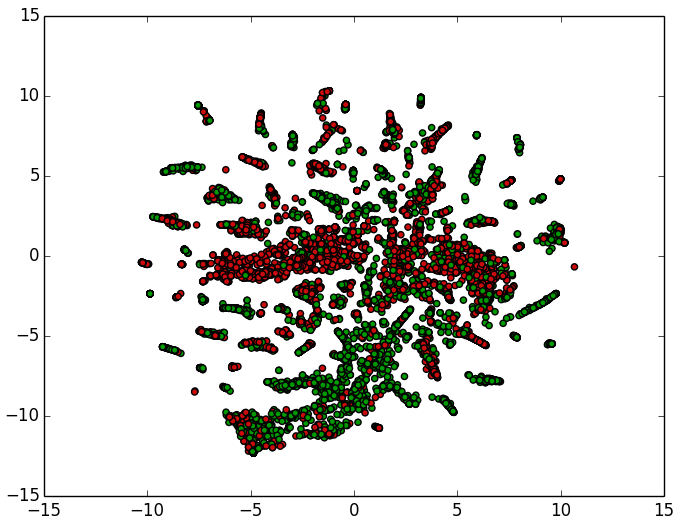}
        \label{fig:youth}
        \subcaption{Youth}
    \end{subfigure}
    \hfill
    \begin{subfigure}[b]{0.49\linewidth}
        \includegraphics[width=\linewidth]{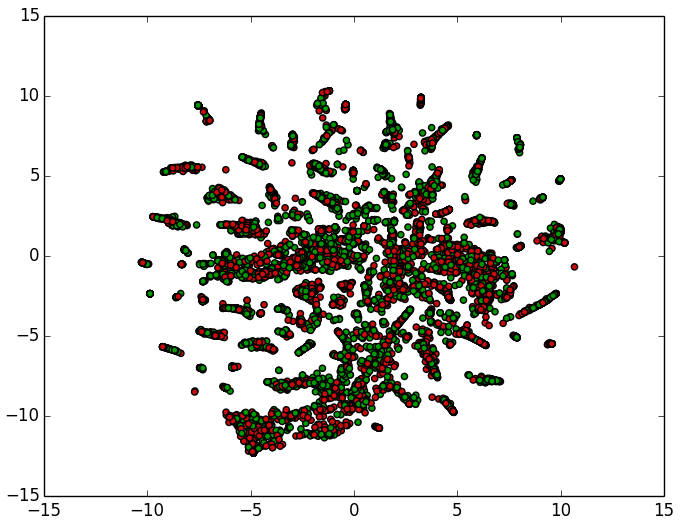}
        \label{fig:softlighting}
        \subcaption{Soft-lighting}
    \end{subfigure}
    \hfill
    \caption{2D representation of fc-7 feature space}
    \label{fig:analysis}
\end{figure}

As mentioned in Section \ref{sec:finetuningmethod}, we use the VGG-Face net as our base for finetuning. Before training the network, we perform some analysis to support the validity of VGG-Face net as our base architecture. Specifically, we perform a forward pass for a set of images from our dataset and obtain activations at the fc-7 layer. We then transform the space of these activations into two dimensions using t-SNE. As evidenced by Figure \ref{fig:analysis}(a), we see clustering even when using the weights directly from VGG-Face net. This gives us confidence that the network is a good candidate for transfer learning. However, clustering does not occur for all of the attributes, as seen in Figure \ref{fig:analysis}(b). This is also expected, since the VGG-Face net learned attributes that were important in distinguishing celebrity faces from each other, and certain attributes like \textit{soft-lighting} may not provide the discriminatory power to do so. After looking at the overall results across the attributes, we decided that we need to backpropagate our gradients into at least some of the layers in the network.

\begin{figure}
  \includegraphics[width=\linewidth]{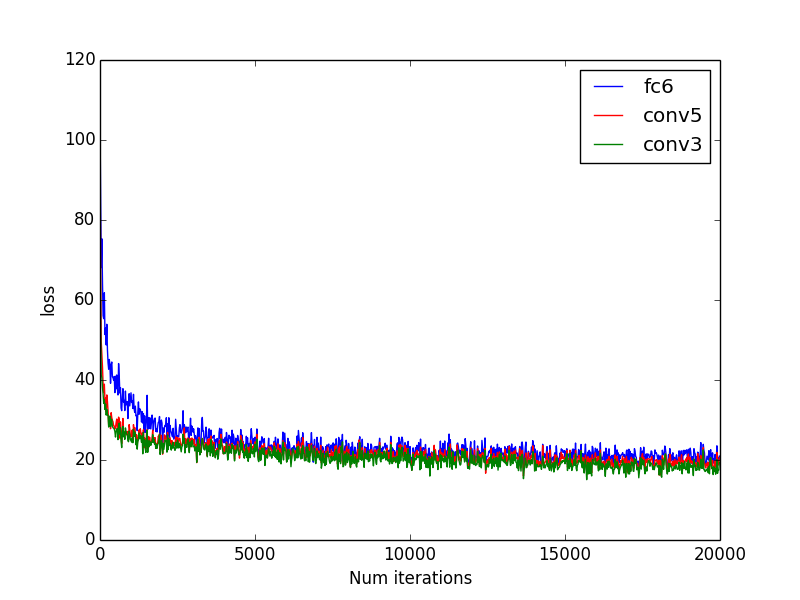}
  \caption{Loss curve after unfreezing various layers in the network}
  \label{fig:loss-hist}
\end{figure}

Next, we tune hyperparameters to get the best results from transfer learning. We try a wide range of learning rates, diagnosing the learning process by looking at the loss curve. We also try various configurations of freezing layers in the network. As we can see from Figure \ref{fig:loss-hist}, backpropagating into more layers in the network helps the learning process, but the final loss is very similar across all trials. We also try different learning rates for the layers based on their distance to the heads. Results were similar to the previous trials, with the network converging to more or less the same loss. 
\subsection{Image Generation Experiments}
\begin{figure}[h!]
    \centering
    \begin{subfigure}[b]{0.24\linewidth}
        \includegraphics[width=\linewidth]{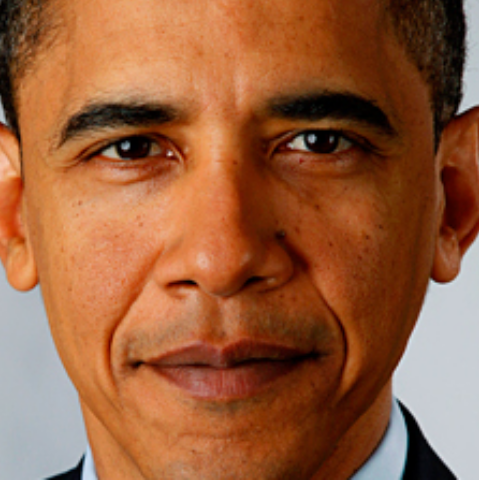}
    \end{subfigure}
    \begin{subfigure}[b]{0.24\linewidth}
        \includegraphics[width=\linewidth]{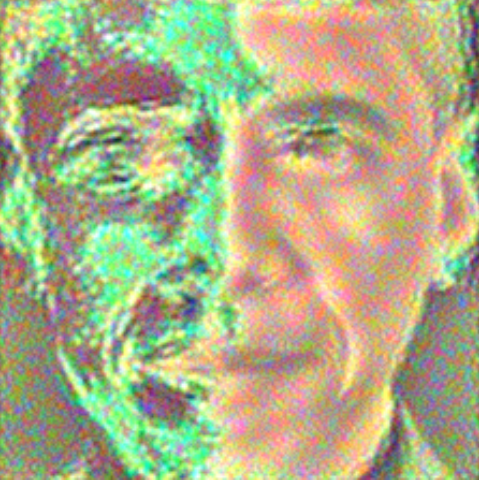}
    \end{subfigure}
    \begin{subfigure}[b]{0.24\linewidth}
        \includegraphics[width=\linewidth]{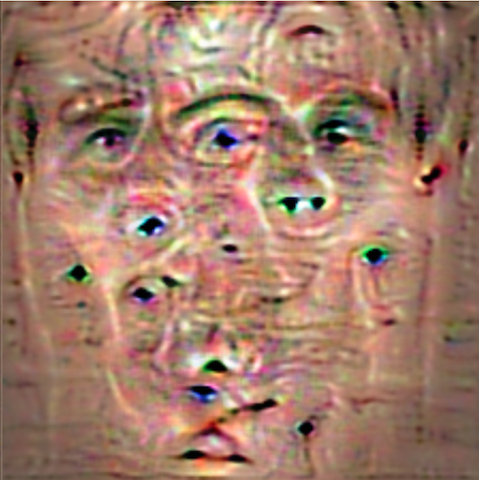}
    \end{subfigure}
    \caption{Target Image, Vanilla Feature Inversion, cGMM Feature Inversion}\label{fig:obamaout}
\end{figure}

We first run several experiments using our baseline class visualization approach. For our primary experiment, we attempt to boost several features that vary in frequency of appearance in our dataset, including \textit{black} (ethnicity) and \textit{black hair}, for which there are fewer images, and \textit{middle-aged}, \textit{smiling}, and \textit{male}, for which the frequency of images is much higher. A resultant image of this experiment is shown in Figure \ref{fig:obamaout}. We see that the final image has multiple facial structures, i.e. multiple sets of noses, eyes, eyebrows, etc. However, the image also contains some semblance of black hair and a facial structure that is more masculine. Our baseline demonstrates that the trained CNN has learned what these general structures in the images look like. 
Because class visualization seems to boost any part of the image whose activations are close to the desired features, we shift our approach to feature inversion. For some image with desired attributes, there exist corresponding activations at an arbitrary layer that we can use to generate the image. To sanity check this idea, we perform an experiment where we use the ground truth activations at a convolutional layer for a Barack Obama image and attempt to reconstruct Obama using these activations. We obtain these activations by performing a forward pass on the image and stopping at the desired convolutional layer. The resultant Barack Obama image is shown in Figure \ref{fig:obamaout}. This result shows that the feature inversion method is a reasonable approach as long as we have access to the "target" activations of an image with selected attributes. As described in our methods, we use a cGMM to estimate these target activations.


\begin{figure}[h]
    \centering
    \hfill
    \begin{subfigure}[b]{0.49\linewidth}
        \includegraphics[width=\linewidth]{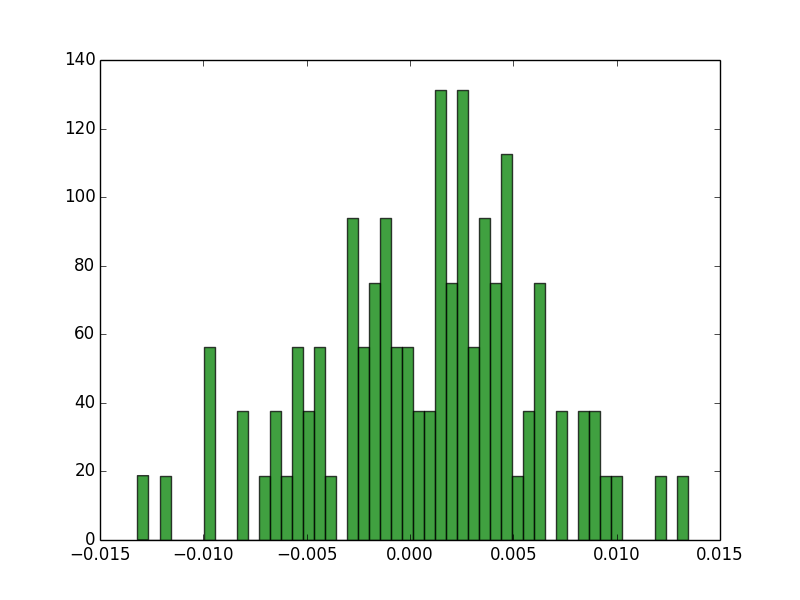}
        \caption{Distribution of element in fc-6 activations for \textit{black} attribute}
        \label{fig:black-1}
    \end{subfigure}
    \hfill
    \begin{subfigure}[b]{0.49\linewidth}
        \includegraphics[width=\linewidth]{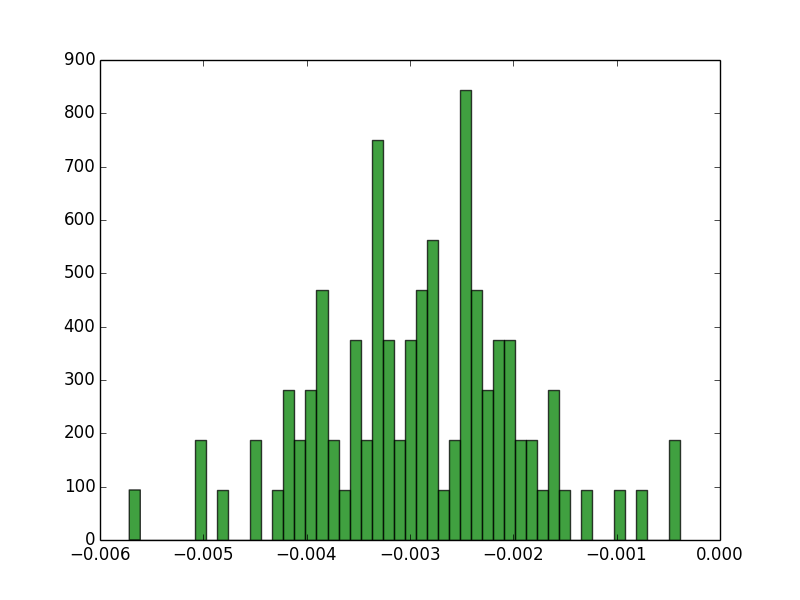}
        \caption{Distribution of element in fc-6 activations for \textit{male} attribute}
        \label{fig:male-1}
    \end{subfigure}
    \hfill
    \caption{2D representation of fc-6 feature space}
    \label{fig:distributions}
\end{figure}

To justify modeling each attribute by an independent multivariate Gaussian, we take the images in set $S_i$ for label $f_i$ and for each image, plot a single element of the activation at the fc-6 layer. We then plot the distribution of that particular element in the activation across all the images for that label $i$. Looking at two distributions representing two elements of the fc-6 activation in Figure \ref{fig:distributions}, we see that the distributions are reasonably close to a normal distribution. In general, after plotting distributions for each element across a sample of images for an arbitrary label, any arbitrary distribution of element $j$ of the fc-6 activation for label $i$ seems to be normal (not all graphs are shown). Thus, because we have some confidence that the true distribution of the target activations is Gaussian, we can proceed with our cGMM formulation. 

Now, we perform experiments to learn the weights in our cGMM. The forward and backward passes of are written from scratch, so we use numerical gradient checking to validate the analytic gradients. To choose the learning rate and regularization parameters, we start by arbitrarily setting the learning rate and regularization parameter to 1e-6 and 1e-5, respectively. Then, we incrementally reduce the learning rate until the loss starts to decrease instead of diverging to infinity. Finally, we use the trained model to generate several faces of differing attributes. 

\section{Results}
\begin{table}[h]
    \centering
    \begin{tabular}{|c|c|c|}
        \hline  & training set & test set \\ \hline
        fc-6 & 0.847 & 0.795 \\
        conv-5 & 0.867  & 0.815 \\ \hline
    \end{tabular}
    \caption{Average classification accuracy freezing parameters beneath layers fc-6 and conv-5}
    \label{tab:finetuningresults}
\end{table}
Our final training parameters for the finetuning are 5e-6 for the learning rate and 50 for the number of epochs. As a form of regularization, we utilize dropout after every layer. The average classification accuracies are shown in Table \ref{tab:finetuningresults}. The classification accuracy for every facial attribute is above $0.5$, while most are around $0.8$ or $0.9$.

\begin{table}[h]
    \centering
    \begin{tabular}{|c|c|}
        \hline person & similarity \% \\ \hline
        Barack Obama & 35 \\
        Clive Owen & 27 \\ 
        Cristiano Ronaldo & 45 \\
        Jared Leto & 52 \\
        Julia Roberts & 42 \\
        Mickey Rourke & 22 \\
        Miley Cyrus & 42 \\
        Nicole Richie & 30 \\
        Ryan Seacrest & 30 \\ \hline
    \end{tabular}
    \caption{Similarity Percentages for Different Faces}
    \label{tab:facesim}
\end{table} 

\begin{figure}[h]
    \centering
    \begin{subfigure}[b]{0.24\linewidth}
        \includegraphics[trim={4.5cm 1.5cm 4.5cm 1.5cm},clip, width=\linewidth]{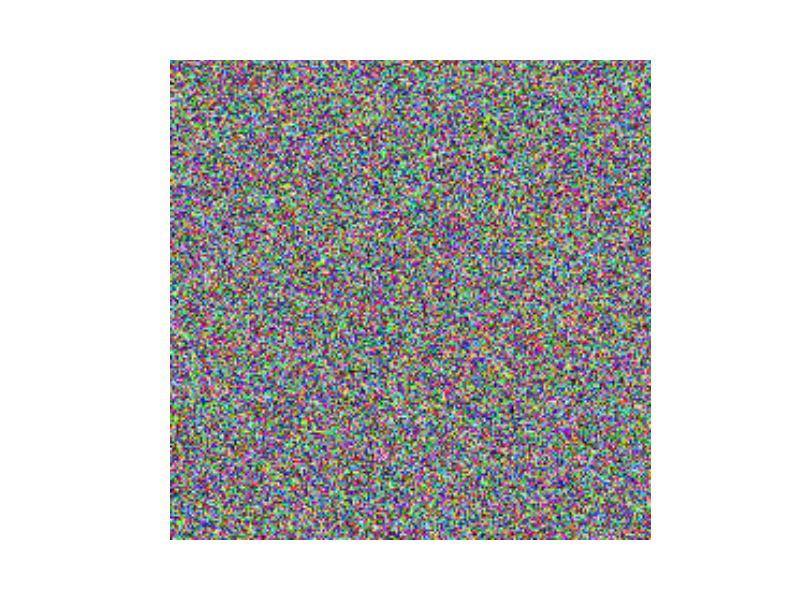}
    \end{subfigure}
    \begin{subfigure}[b]{0.24\linewidth}
        \includegraphics[trim={4.5cm 1.5cm 4.5cm 1.5cm},clip, width=\linewidth]{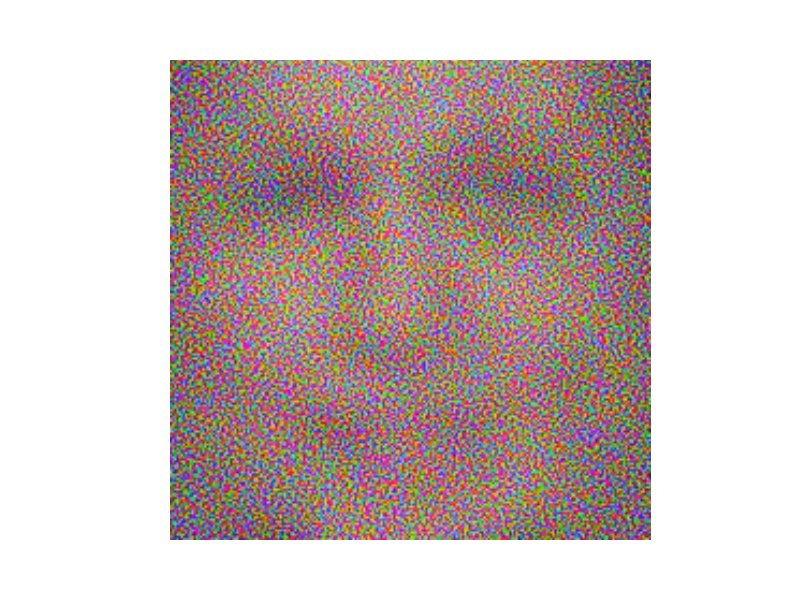}
    \end{subfigure}
    \begin{subfigure}[b]{0.24\linewidth}
        \includegraphics[trim={4.5cm 1.5cm 4.5cm 1.5cm},clip, width=\linewidth]{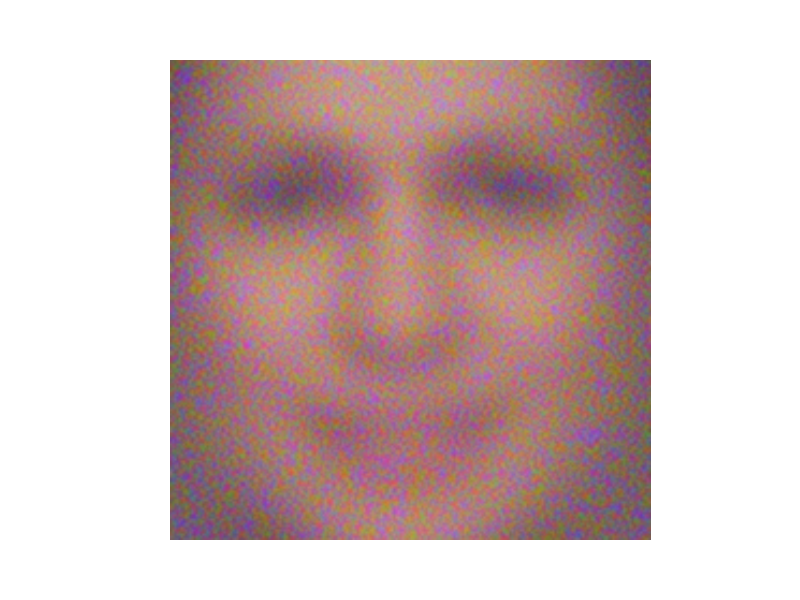}
    \end{subfigure}
    \begin{subfigure}[b]{0.24\linewidth}
        \includegraphics[trim={4.5cm 1.5cm 4.5cm 1.5cm},clip, width=\linewidth]{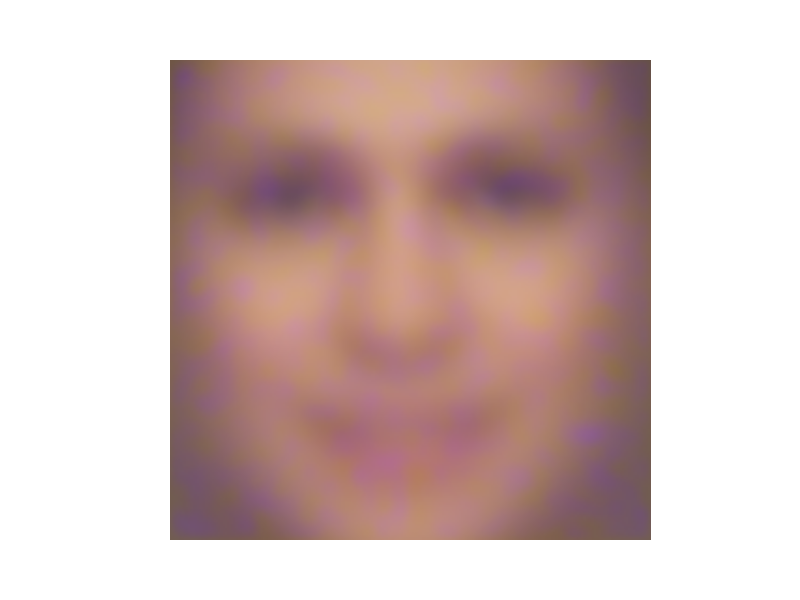}
    \end{subfigure}
    \caption{Cristiano Ronaldo generated image}
    \label{fig:ronaldo}
\end{figure}

For our cGMM training, our final learning rate and regularization parameter are 1e-11 and 1e-5, respectively. We generate faces for sets of attributes that are defined by several images in the test set (i.e. we plug in these target attributes into our system and compare the image we generate with the ground truth image). One example of a generated image is in Figure \ref{fig:ronaldo}. To measure how "reasonable" our faces are, we use a similarity metric implemented by the PicTriev software, which uses common facial features to measure similarity \cite{pict}. This quantitative evaluation of our faces is shown in Table \ref{tab:facesim}.

\section{Discussion}
Although our classification accuracies are generally quite high, there are some facial attributes such as lighting (e.g., \textit{soft lighting}, \textit{harsh lighting}) and photo type (e.g., \textit{color photo}, \textit{posed photo}) that have low accuracies (roughly 55-60\%). We do not attempt to improve these attributes, since they are not as important as other, more face-related characteristics, such as \textit{nose shape} and \textit{forehead visibility}, that have high classification accuracies.

The most obvious result from our generation experiments is that each resultant image is very close to the mean image of our training set.
Despite this, the fact that we successfully generate a coherent face from random noise indicates that our feature inversion process could work if the activations for each attribute were more discriminative.

We hypothesize several reasons why our activations are not discriminative enough. One possibility is that the distributions for different attributes are very similar and clustered around the mean image in this abstract space.
Another possibility is that the distributions lie in different manifolds in this space, but that they are not normal yet roughly centered around the mean-image.
This means that the act of averaging these activations and approximating them as Gaussian blurs the distinctions between these distributions.

This bring us to the assumptions about the feature activations upon which the cGMM approach relies.
Our first assumption is that the distribution of activations for each attribute in a given layer is indeed Gaussian.
Additionally, we assume that these Gaussians are independent. 
This is perhaps the least justified assumption the model makes, because clearly many attributes are correlated, such as \textit{middle-aged} and \textit{white hair}.
Despite this, we hope that our model approximates the true distribution sufficiently for our purposes.
We also assume that the variables in the multivariate Gaussian are themselves distributed normally and are independent.
We believe this assumption is valid because we examined several of these distributions for individual variables as described in Section 5.

One way to improve our cGMM technique would be to estimate the target activations of a desired image by taking a weighted sum of \emph{samples} from the Gaussians instead of a weighted sum of the \emph{means} of the Gaussians.
We opted for the latter approach because it was far more computationally efficient, but the former approach could allow for more variation from the mean image.

\section{Conclusion \& Future Work}
In this work, we have trained an architecture that classifies facial attributes of images with high accuracy. We have also shown the effectiveness of the custom Gaussian Mixture Model in generating well-formed faces from random noise. 
Although the faces themselves were not discriminative enough for the desired features, we posit that we can train a model that generates discriminative faces if our dataset contained more diversity across the attributes.

Alternatively, we might implement a variational autoencoder structure as in \cite{gregor2015draw} and compare the results with those of the cGMM. We expect that since the variational autoencoder has a higher capacity, it might give more discriminative faces but would also take longer to train. More efficient generative models open up exciting possibilities in other applications, including medicine and law enforcement. 

\newpage
\nocite{*}
{\small
\bibliographystyle{ieee}
\bibliography{references}

\begin{thebibliography}{10}\itemsep=-1pt

\bibitem{pict}
AppliedDevice.
\newblock Pictriev: searching faces on the web, 2010.
\newblock [Online; accessed 2016-03-14].

\bibitem{bengio2013generalized}
Y.~Bengio, L.~Yao, G.~Alain, and P.~Vincent.
\newblock Generalized denoising auto-encoders as generative models.
\newblock In {\em Advances in Neural Information Processing Systems}, pages
  899--907, 2013.

\bibitem{gauthier2014conditional}
J.~Gauthier.
\newblock Conditional generative adversarial nets for convolutional face
  generation.
\newblock {\em Class Project for Stanford CS231N: Convolutional Neural Networks
  for Visual Recognition, Winter semester}, 2014, 2014.

\bibitem{goodfellow2014generative}
I.~Goodfellow, J.~Pouget-Abadie, M.~Mirza, B.~Xu, D.~Warde-Farley, S.~Ozair,
  A.~Courville, and Y.~Bengio.
\newblock Generative adversarial nets.
\newblock In {\em Advances in Neural Information Processing Systems}, pages
  2672--2680, 2014.

\bibitem{gregor2015draw}
K.~Gregor, I.~Danihelka, A.~Graves, and D.~Wierstra.
\newblock Draw: A recurrent neural network for image generation.
\newblock {\em arXiv preprint arXiv:1502.04623}, 2015.

\bibitem{jia2014caffe}
Y.~Jia, E.~Shelhamer, J.~Donahue, S.~Karayev, J.~Long, R.~Girshick,
  S.~Guadarrama, and T.~Darrell.
\newblock Caffe: Convolutional architecture for fast feature embedding.
\newblock {\em arXiv preprint arXiv:1408.5093}, 2014.

\bibitem{}
E.~Jones, T.~Oliphant, P.~Peterson, et~al.
\newblock {SciPy}: Open source scientific tools for {Python}, 2001--.
\newblock [Online; accessed 2016-03-14].

\bibitem{dataset}
N.~Kumar, A.~C. Berg, P.~N. Belhumeur, and S.~K. Nayar.
\newblock {A}ttribute and {S}imile {C}lassifiers for {F}ace {V}erification.
\newblock In {\em IEEE International Conference on Computer Vision (ICCV)}, Oct
  2009.

\bibitem{deep2016tutorial}
L.~Lab.
\newblock Denoising autoencoders (da).
\newblock 2016.

\bibitem{marc2007efficient}
C.~P. Marc’Aurelio~Ranzato, S.~Chopra, and Y.~LeCun.
\newblock Efficient learning of sparse representations with an energy-based
  model.
\newblock In {\em Proceedings of NIPS}, 2007.

\bibitem{vggfacenet}
A.~Z. O.~M.~Parkhi, A.~Vedaldi.
\newblock {D}eep {F}ace {R}ecognition.
\newblock In {\em British Machine Vision Conference (BMVC)}, 2015.

\bibitem{salakhutdinov2009learning}
R.~Salakhutdinov.
\newblock {\em Learning deep generative models}.
\newblock PhD thesis, University of Toronto, 2009.

\bibitem{salakhutdinov2009deep}
R.~Salakhutdinov and G.~E. Hinton.
\newblock Deep boltzmann machines.
\newblock In {\em International conference on artificial intelligence and
  statistics}, pages 448--455, 2009.

\bibitem{NIPS20124683}
N.~Srivastava and R.~R. Salakhutdinov.
\newblock Multimodal learning with deep boltzmann machines.
\newblock In F.~Pereira, C.~J.~C. Burges, L.~Bottou, and K.~Q. Weinberger,
  editors, {\em Advances in Neural Information Processing Systems 25}, pages
  2222--2230. Curran Associates, Inc., 2012.

\bibitem{vincent2011connection}
P.~Vincent.
\newblock A connection between score matching and denoising autoencoders.
\newblock {\em Neural computation}, 23(7):1661--1674, 2011.

\bibitem{vincent2010stacked}
P.~Vincent, H.~Larochelle, I.~Lajoie, Y.~Bengio, and P.-A. Manzagol.
\newblock Stacked denoising autoencoders: Learning useful representations in a
  deep network with a local denoising criterion.
\newblock {\em The Journal of Machine Learning Research}, 11:3371--3408, 2010.

\end{thebibliography}
}

\end{document}